\documentclass[preprint]{sig-alternate-05-2015}

\usepackage{pdf14}
\usepackage[breaklinks=true,bookmarks=false,colorlinks=true]{hyperref}
\usepackage{subfig}
\usepackage{graphicx}
\usepackage{multirow}
\usepackage{multicol}
\usepackage{makecell}
\usepackage{balance}

\makeatletter
\def\@copyrightspace{\relax}
\makeatother
\begin{document}

\title{SwiDeN : Convolutional Neural Networks For Depiction Invariant Object Recognition}

\numberofauthors{1}
\author{\alignauthor Ravi Kiran Sarvadevabhatla, Shiv Surya\titlenote{Equal contributor as the first author}, Srinivas S S Kruthiventi , and\\ Venkatesh Babu R. \\
\affaddr{Video Analytics Lab, Department of Computational and Data Sciences, Indian Institute of Science} \\ \affaddr{Bangalore, India} \\
\affaddr{ravikiran@grads.cds.iisc.ac.in, shiv.surya314@gmail.com, kssaisrinivas@gmail.com, venky@cds.iisc.ac.in}
}

\maketitle
\begin{abstract}
Current state of the art object  recognition architectures achieve impressive performance but are typically specialized for a single depictive style (e.g. photos only, sketches only). In this paper, we present SwiDeN: our Convolutional Neural Network (CNN) architecture which recognizes objects regardless of how they are visually depicted (line drawing, realistic shaded drawing, photograph etc.). In SwiDeN, we utilize a novel `deep' depictive style-based switching mechanism which appropriately addresses the depiction-specific and depiction-invariant aspects of the problem. We compare SwiDeN with  alternative architectures and prior work on a $50$-category Photo-Art dataset containing objects depicted in multiple styles. Experimental results show that SwiDeN outperforms other approaches for the depiction-invariant object recognition problem. 
\end{abstract}
\keywords{object category recognition, convolutional neural networks, deep learning, depiction-invariance}

\section{Introduction}

\begin{figure}[!tbp]
    \centering    
		\includegraphics[width=1.0\linewidth]{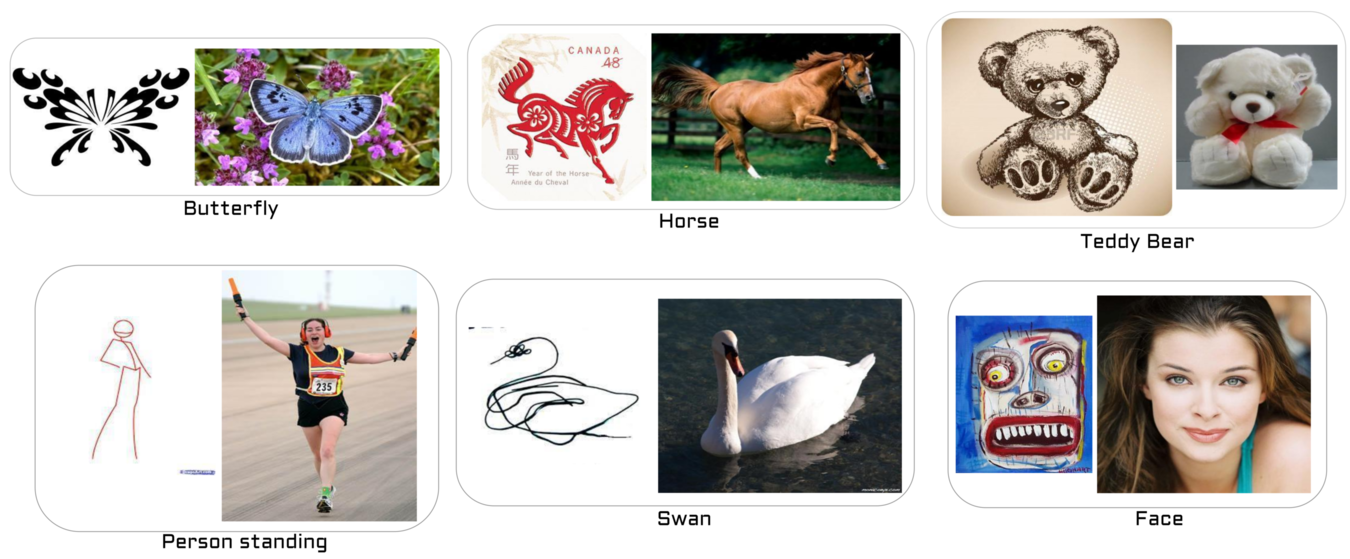}
		\caption{Sample images from the Photo-Art-50 dataset grouped by category. For each category, one image each from `Art'(left in the pair) and `Photo' depictive style are shown. Given the extreme changes in appearance, recognizing such images regardless of depiction is extremely challenging.} 
		\label{fig:datasetpic}    
\end{figure}

Depiction-invariant object recognition is the ability to determine an object's category regardless of how the object is visually depicted (line drawing, realistic shaded drawing, photograph etc.). Given the varying level of abstraction and complexity in depiction (See Figure \ref{fig:datasetpic}), this is a challenging task. Human beings easily accomplish depiction-invariant recognition but machine-based systems are nowhere close to a similar level of performance. Current state-of-the-art object recognition architectures do achieve good performance but they are specialized for a single depiction style (e.g. photos~\cite{he2015deep}, sketches~\cite{yang2015deep}). Therefore,  designing architectures which recognize objects regardless of depiction style can facilitate progress towards matching human-level abilities. Moreover, the associated performance scores can also aid in quantitatively determining the semantic gap between human and machine capabilities~\cite{Ullman08032016}. 

Surprisingly, not much work exists for depiction-invariant object recognition. To address this gap, we propose a Convolutional Neural Network (CNN) architecture for depiction-invariant object category recognition which we call SwiDeN (Section \ref{sec:framework}). A novel aspect of our architecture is a `deep' dynamic switching mechanism between two parallel CNN sub-architectures (Section \ref{switch}). Our switch-based design not only reduces the overall burden of the generalized object recognition task but also enables the system to address depiction-specific and depiction-invariant aspects of the problem. We compare our approach with baselines,  alternative architectures (Section \ref{sec:comparisonarch}) and previous work on a $50$-category Photo-Art dataset containing multiple depictions of objects (Section \ref{sec:experiments}). Experimental results show that our architecture outperforms other architectures, especially for non-photo object depictions (Section \ref{sec:results}). 

\begin{figure*}[ht]
	\mbox{\subfloat[Baseline] { \includegraphics[width =2.65cm, height=9.0cm]{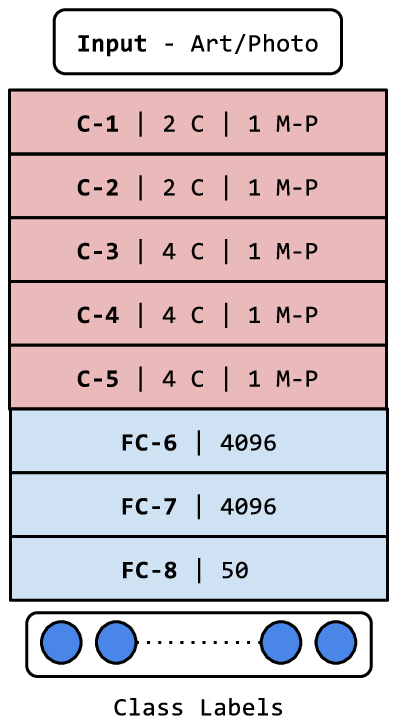}}}\hspace{0.3cm}%
	\mbox{\subfloat[Gradient Reversal Network(GRN)] { \includegraphics[width =5cm, height=9.0cm]{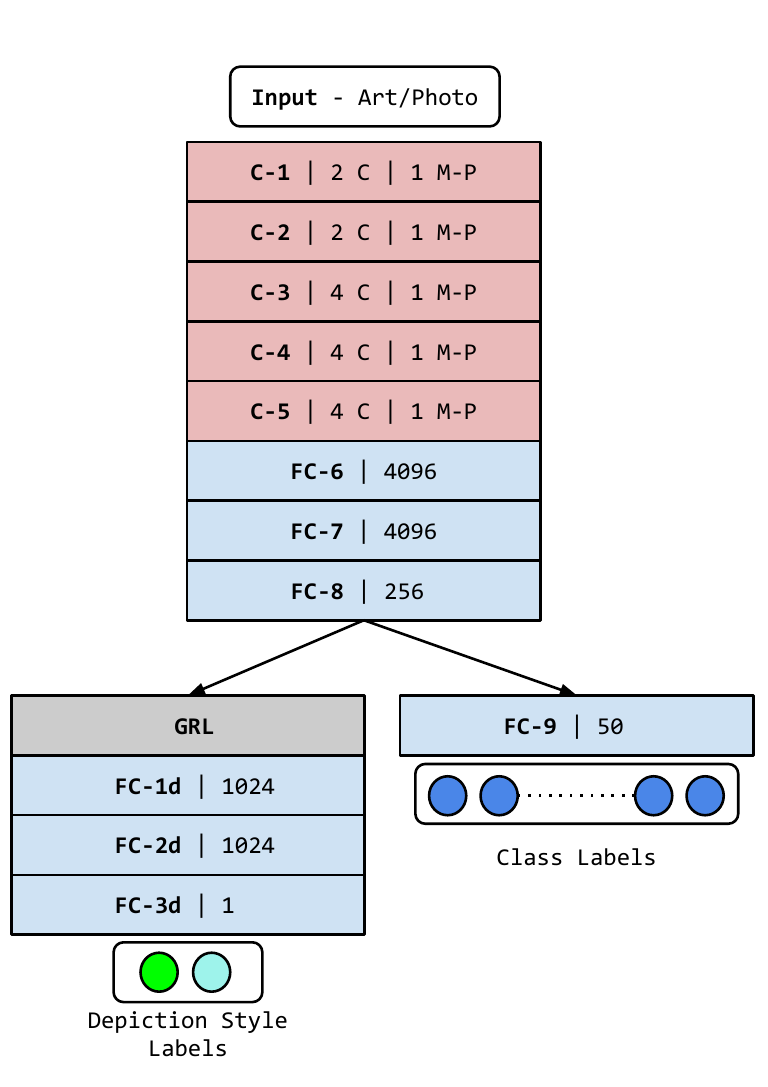}}}\hspace{0.3cm}%
	\mbox{\subfloat[Switching Deep Networks(SwiDeN)] { \includegraphics[width =10.0cm, height=10.0cm]{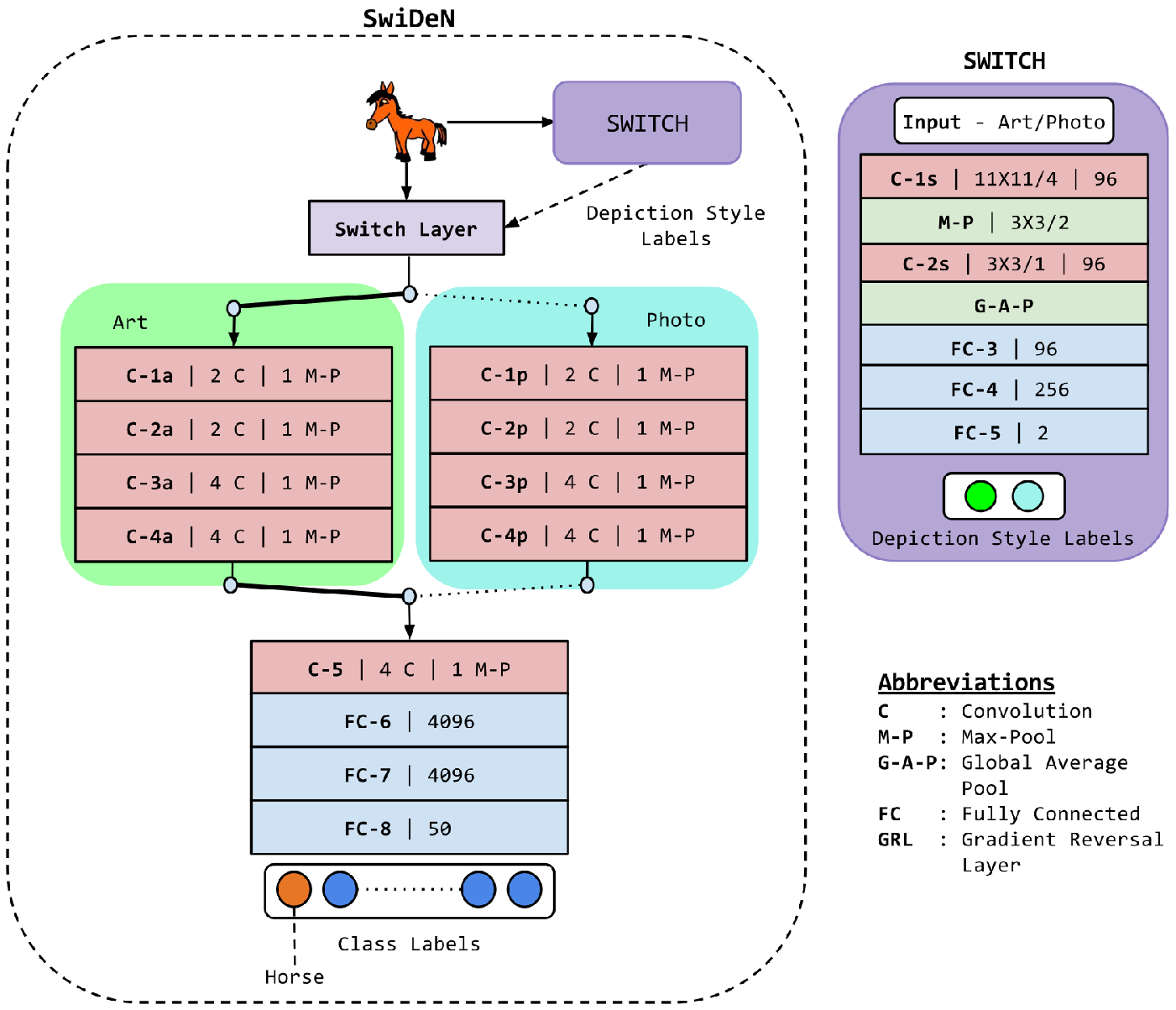}}}\hspace{0.3cm}%
\caption{Our proposed architecture SwiDeN is shown in \ref{fig:3imgs}(c). The depictive style of the  cartoon-ish horse image is  determined as `Art' by  \textsc{Switch} (purple block). An associated switch layer relays it to the `Art' sub-network (green block). The latter's output is passed via a series of shared layers and finally, a softmax classifier generates the label \texttt{Horse}. Figure  \ref{fig:3imgs}(a) is the baseline architecture. Figure \ref{fig:3imgs}(b) (GRN) is a modification of architecture proposed by Ganin et al~\cite{ganin2015unsupervised}. VGG-19~\cite{Simonyan14c} is used as the base network for all architectures.}
\label{fig:3imgs}
\end{figure*}

\section{Related Work}
\label{sec:relatedwork}

Object class (category) recognition, albeit restricted to photographic depictions, has been studied extensively by researchers~\cite{fergus2003object,fidler2007towards,shotton2008multiscale}.
However, little previous work exists for truly general multi- depiction object recognition. Wu et al.~\cite{wu2014learning} construct  multi-attribute part-graphs for object categories and use graph matching for classification on the same dataset we use. However, their evaluation procedure, also used by Cai et al.~\cite{cai2015cross,cai2015beyond}, induces an unreasonable amount of category bias which makes comparison difficult. We present an alternative evaluation procedure which is more principled (See Section \ref{datasets}). Xiao et al.~\cite{Xiao20111023}
 present a graph-based object modelling approach and evaluate it on $10$ augmented classes of Caltech-256. Shrivastava et al~\cite{shrivastava-sa11} utilize a depiction-invariant method for image matching. Domain adaption approaches have been also been tried~\cite{cai2015beyond}. However, when the domain-specific identifiers (e.g. target domain labels) are available as in our case, a domain-adaptation procedure unnecessarily makes the overall problem harder since the objective in domain-adaptation is typically to ``forget" the source domain. 

All the approaches mentioned above utilize multiple hand-crafted modules in the recognition pipeline. To the best of our knowledge, ours is the first end-to-end deep learning approach for depiction-invariant recognition of object categories.

\section{Our framework}
\label{sec:framework}

\subsection{Motivation}
\label{sec:motivation}

Instead of learning from scratch, a common paradigm is to utilize pre-trained CNNs as a starting point while constructing deep networks of interest. We follow a similar paradigm in our approach. 

In an effort to represent the sheer variety seen in image content, the convolutional layers in a CNN typically contain a large number of learnable filters. However, the filters are only sufficient to the extent that the depiction style remains unchanged (e.g. photographs). To accommodate the increase in variety when images from additional depiction styles need to be recognized, a na\"{\i}ve strategy would be to add additional learnable filters for each convolution layer of a pre-trained network\footnote{In this case, the network could be one pre-trained for a particular depiction style (e.g. photographs).} and perform fine-tuning. However, this strategy results in an unbalanced learning regime since convolutional layers now contain a mixture of learnt and non-learnt filters. In addition, the added filters necessitate an ad-hoc grouping of filter layers to ensure operational consistency which further complicates the overall framework. 

An alternative design would be to learn the filters for each depictive style separately. In this design, a set of shallow layer sub-networks exist for each depictive style (see Figure \ref{fig:3imgs}(c)). Since our final objective is to achieve depiction-invariant recognition, we require our network to learn a depiction-invariant feature representation. This is achieved by having a final set of layers. To serve as a relay mechanism between the initial depiction-specific sub-network branches and the shared, deeper depiction-invariant fully-connected layers, we employ a custom-designed ``switch" (Section \ref{switch}). The switch is trained such that given an image, it determines its depictive style and selects the corresponding depiction-specific sub-network for processing the image. The output of this sub-network is then processed by the depiction-invariant layers of the network. The network culminates in a typical softmax-based classification layer which determines the image category, regardless of its depictive style (Figure \ref{fig:3imgs}(c)). 

Next, we describe the depiction style-based switching mechanism. Subsequently, we delve into the architectural details of the main network pipeline which we dub SwiDeN (Switching Deep Network).

\subsubsection{\textsc{Switch}}
\label{switch}

To realize the switching mechanism mentioned in Section \ref{sec:motivation}, we design and train a switch network (see Figure \ref{fig:3imgs} (c)), henceforth referred to as \textsc{Switch}, that determines the depiction style of the input image and passes the image to corresponding depiction sub-network (Photo or Art). The \textsc{Switch} has two convolution layers which capture depiction-discriminative features such as edges, textures, corners, colors and their  conjunctions~\cite{zeiler2014visualizing}.  The first convolution layer is initialized from AlexNet~\cite{krizhevsky2012imagenet}. The features from the first layer are max pooled while the features from the second convolution layer are average pooled globally~\cite{lin2013network}. The pooled features are processed by two fully connected layers and passed to a classifier layer which determines the depiction style of the input image as `Art' or `Photo'. For better generalization, we use dropout for fully connected layers with $0.5$ as the dropout value. We trained \textsc{Switch} using stochastic gradient descent (SGD)~\cite{krizhevsky2012imagenet} with a base learning rate $\alpha=10^{-2}$ and momentum $\mu=0.9$. Overall, \textsc{Switch} achieves an average accuracy of $83.7\%$ ($80.6\%$ for `Art' and $86.8\%$ for `Photo'). 

\textsc{Switch}'s inability to achieve $100\%$ accuracy can be attributed to the fact that some photo images have a predominantly artistic quality and vice-versa (see supplementary material). While this may seem like a liability, in practice, all we require is that \textsc{Switch} achieve a reasonably high accuracy which ensures an overall burden reduction for the filter learning process.

\subsubsection{\textbf{Switching Deep Network (SwiDeN)}}
\label{swiden}

The initial portion of SwiDeN consists of two separate sub-networks, one each for photo and art depiction style. During training, \textsc{Switch} (Section \ref{switch}) selects the sub-network branch through which the input image is passed in the forward pass and ensures that the corresponding network loss is backpropagated through the branch selected during the forward pass. The layers after \textsc{Switch} are shared layers, designed to learn   depiction-style invariant representations. 

For our problem, we build SwiDeN using VGG-19 deep network~\cite{Simonyan14c} layers. We select a subset of initial convolutional layers of VGG-19 and utilize them as the sub-networks for each depictive style. The rest of the VGG-19 layers (except the final classification layer) are used as the shared layers of SwiDeN. Figure \ref{fig:3imgs}(c) illustrates a SwiDeN architecture where the first four convolutional layers of VGG-19 are used for the depictive style sub-networks (\texttt{C1a} - \texttt{C4a} for `Art' and \texttt{C1p} - \texttt{C4p} for `Photo') and the rest of VGG-19 layers  (\texttt{C-5,FC-6,FC-7}) form the shared portion. 

In our experiments, we systematically examined the effect on recognition performance when the first $k, 1\le k \le 5$ convolutional layers of VGG-19 are used as depiction-style sub-networks  (Section \ref{sec:results}). For the rest of the paper, we refer to the corresponding architectures as \textbf{C1-S,C2-S,C3-S,C4-S} and \textbf{C5-S}. Thus, the architecture in Figure \ref{fig:3imgs}(c) is \textbf{C4-S}. 

\section{Experiments}
\label{sec:experiments}
 
\subsection{Dataset}
\label{datasets}
    We evaluate the classification performance on the Photo-Art-50 dataset~\cite{cai2015cross}. This dataset contains $50$ classes and $90$ to $138$ images in each class with approximately half photo and half art images. The authors also provide train-test splits for comparative evaluation. However, the splits are unbalanced and do not include a validation split, thus inducing significant class bias during evaluation. To avoid this issue, we  create our own train, validation and test splits. We create five random splits, each containing $60$ images from each category for training ($30$ art and $30$ photo) and $20$ images for testing ($10$ art and $10$ photo). The remaining images from each category are used for validation. We augment the dataset by taking 5 crops of size $224\times 224$ (four corner crops and the center crop) after rescaling the smallest side of the image to $256$. For training images, the center crop alone is centered around the bounding box. Multiple objects of same class in a single image are ignored. We plan to release our balanced splits to the public.  

\subsection{Comparison architectures}
\label{sec:comparisonarch}

\subsubsection{Baseline}
\label{sec:baseline}
As a natural baseline, we fine-tune VGG-19 using the training data for the $50$ classes described in Section \ref{datasets}. For training, we used a stochastic gradient descent (SGD) method with a base learning rate $\alpha=10^{-5}$ and momentum $\mu=0.9$ to learn the weights. The learning rate was stepped down by a factor of $10$ when the validation accuracy plateaued.

\subsubsection{Gradient Reversal Network}
\label{grn}

Ganin et al.~\cite{ganin2015unsupervised} propose a deep network-based domain-adaptation framework. The authors aim to maximize the target domain accuracy by simultaneously minimizing the target-domain label loss function and maximizing the loss for domain type (target or source) classification. To achieve this, they introduce a gradient reversal layer which not only assists domain-adaptation but also helps learn a domain-invariant representation (FC-8 in Figure \ref{fig:3imgs}(b)). Intrigued by this domain-invariance feature, we wished to examine the architecture's suitability for our cross-depiction problem by viewing depictive styles as domains. However, in their original formulation, Ganin et al. maximize the accuracy for a single domain (depictive style). Therefore, we modify their formulation such that the overall network loss for \textit{both} the domains (`Art' and `Photo') is minimized. In addition, we replace Alexnet used by Ganin et al. with VGG-19. For the rest of the paper, we shall refer to this modified formulation as Gradient Reversal Network (GRN). 

We initialize GRN with VGG-19 model weights and performed training using SGD with base learning rate of $\alpha=10^{-5}$ and momentum $\mu=0.9$. A uniform learning rate was maintained throughout training. For the gradient reversal layer's scaling factor $\lambda$ (see \cite{ganin2015unsupervised} for details), we tried values of $1,2,3,5,10$ and  found that $\lambda=2$ gave the best result.  

\subsubsection{SwiDeN: training}
\label{swtr}

The same training procedure and hyperparameters as in the baseline were used for training SwiDeN architectures \textbf{C1-S,C2-S}$\ldots$\textbf{C5-S} (Section \ref{swiden}) with the exception of the the learning rates for the depictive style sub-networks. For the `Art' sub-network, we used a learning rate scaled by a factor of $4$ since the base network (VGG) is primarily trained for non-Art images. The learning rate was stepped down by a factor of $10$ when the validation accuracy plateaued.

\subsection{Evaluation}
For evaluation, we determined the final label by  pooling the results for five crops of the test image (four corner crops and one center crop) for all the architectures.

\subsection{Implementation}
 We used Caffe~\cite{jia2014caffe} for all experiments on the baseline. For SwiDeN, we integrated the switch layer from a branch of Caffe~\cite{caffe-dev} into the master branch~\cite{caffe-master} and customized it for our experiments involving SwiDeN. For experiments on GRN, we used the Caffe version provided by  Ganin et al.~\cite{ganin2015unsupervised}. 
 
\section{Results}
\label{sec:results}

\renewcommand{\arraystretch}{1.5}
\begin{table}[!tbp]
\centering
\footnotesize
\begin{tabular}{|c|c|c|c|}
\hline
\textsc{Arch.} & \thead{\textsc{Overall} \\ \textsc{Acc.}} & \thead{\textsc{Art} \\ \textsc{Acc.}} & \thead{\textsc{Photo} \\ \textsc{Acc.}} \\
\hline
\hline 
 Baseline & 93.80\% & 89.80\% & \textbf{97.80\%} \\
\hline
GRN & 92.64\% & 88.52\%  & 96.76\% \\
\hline
\textbf{SwiDeN(Ours)} & \textbf{94.42\%} & \textbf{91.12\%} & 97.72\% \\
\hline
\end{tabular}
\caption{Classification accuracy for different architectures.}
\label{tab:accuracy}
\end{table}

For each architecture (baseline, GRN and SwiDeN (\textbf{C4-S}), we computed the average test set accuracy across all the classes and all the splits. We collate the results into three groups -- accuracy regardless of depictive style (`Overall') and style-wise accuracies for `Photo' (i.e. accuracy on photographic test images only) and `Art'). The results can be seen in seen in Table \ref{tab:accuracy}. Our SwiDeN architecture outperforms the other two architectures overall and for `Art' while remaining competitive for `Photo'. In SwiDeN, the depiction-style \textsc{Switch} guided sub-network learning reduces the overall burden for the deeper shared layers in learning a robust depiction-invariant representation, which in turn contributes to SwiDeN's performance. 

GRN performs worse than the baseline and SwiDeN. Similar to SwiDeN, GRN also utilizes feedback from a depiction-style classifier. However, the feedback is provided coarsely and indirectly (in terms of loss). Moreover, the feedback is provided at a layer situated deep in the network. This hinders the fine-tuning of shallower (convolutional) filters to learn `Art'-specific filters, thus affecting the performance on `Art' in particular and overall performance in general.

We also observe that the baseline performs slightly better than other architectures for `Photo' style. This is to be expected since the original filters are highly-tuned for photos. However, its performance for `Art' is relatively lower compared to SwiDeN. This shows that the complexity involved in cross-depiction recognition cannot be addressed merely by employing typical transfer learning approaches such as fine-tuning.

In spite of the class-bias induced by the splits provided by Cai et al.~\cite{cai2015cross}, we compared the performance of our \textbf{C4-S} SwiDeN architecture against that of the multi-attribute part-graph model proposed by Wu et al.~\cite{wu2014learning}. To aid training, we augment the training set by performing RGB jittering, horizontal flip on all images and morphological operations for `Art' images. As Table \ref{tab:accuracy2} shows, SwiDeN achieves state-of-the-art results , outperforming the result of Wu et al.~\cite{wu2014learning} overall and for `Photo' images while remaining competitive for `Art'.

Table \ref{tab:accuracy3} summarizes the performance of different SwiDeN architectures. As can be seen, \textbf{C4-S} outperforms other SwiDeN architectures. As an interesting observation, the trends in overall accuracy and `Art' accuracy  as the depth of depictive-style sub-networks increases resemble the patterns observed by Yosinski et al.\cite{yosinski2014transferable} for deep networks but in the context of transfer learning. 

\renewcommand{\arraystretch}{1.5}
\begin{table}[!tbp]
\centering
\footnotesize
\begin{tabular}{|c|c|c|c|}
\hline
\textsc{Arch.} & \thead{\textsc{Overall} \\ \textsc{Acc.}} & \thead{\textsc{Art} \\ \textsc{Acc.}} & \thead{\textsc{Photo} \\ \textsc{Acc.}} \\
\hline
\hline 
Wu et al.\cite{wu2014learning} & 89.67\% & \textbf{89.06}\%  & 90.29\% \\
\hline
\textbf{SwiDeN (Ours)} & \textbf{93.02\%} & 88.47\% & \textbf{97.56\%} \\
\hline
\end{tabular}
\caption{Classification accuracy on train-test splits by Cai et al.~\cite{cai2015cross}.}
\label{tab:accuracy2}
\end{table}

\renewcommand{\arraystretch}{1.5}
\begin{table}[ht]
\centering
\footnotesize
\begin{tabular}{|c|c|c|c|}
\hline
\textsc{ SwiDeN Arch.} & \thead{\textsc{Overall} \\ \textsc{Acc.}} & \thead{\textsc{Art} \\ \textsc{Acc.}} & \thead{\textsc{Photo} \\ \textsc{Acc.}} \\
\hline
\hline 
\textbf{C1-S} & 94.22\% & 90.44\% & \textbf{98.00\%} \\
\hline
\textbf{C2-S} & 94.36\% & 90.8\%  & 97.92\% \\
 \hline
\textbf{C3-S} & 93.96\% & 90.4\% & 97.52\% \\
\hline
\textbf{C4-S} & \textbf{94.42\%} & \textbf{91.12\%} & 97.72\% \\
\hline
\textbf{C5-S} & 92.64\% & 88.52\%  & 96.76\% \\ 
\hline
\end{tabular}
\caption{Classification accuracy for different SwiDeN architectures \textbf{C1-S--C5-S}(see Section \ref{swiden}).}
\label{tab:accuracy3}
\end{table}

\section{Conclusion}

In this paper, we have described SwiDeN, our end-to-end deep learning framework for recognizing objects regardless of depiction. A key aspect of SwiDeN is the `deep' depictive style-based switching mechanism which judiciously addresses  depiction-specific and depiction-invariant aspects of the problem. Addressing these aspects enables us to achieve state-of-the-art results on a challenging dataset containing `Photo' and `Art' style object depictions. In future, we plan to explore unsupervised network learning approaches. Our code and pre-trained models can be accessed at \url{https://github.com/val-iisc/swiden}.

\bibliographystyle{abbrv}

\balancecolumns

\end{document}